\documentclass[letterpaper, 10 pt, conference]{ieeeconf}  

\IEEEoverridecommandlockouts                              

\usepackage{graphics} 
\usepackage{epsfig} 
\usepackage{mathptmx} 
\usepackage{times} 
\usepackage{amsmath} 
\usepackage{amssymb}  
\usepackage{newtxmath}  
\usepackage{algorithm}
\usepackage{algorithmic}
\usepackage{optidef}
\usepackage{float}
\usepackage{graphicx}
\usepackage{multirow}
\usepackage{multicol}
\usepackage{booktabs}
\usepackage{subcaption}
\usepackage{placeins}
\usepackage{siunitx}
\usepackage{url}
\usepackage{xcolor}
\usepackage{placeins} 
\usepackage[font=normalsize,labelfont=bf]{caption}  
\usepackage[switch]{lineno}  

\newcommand{\worldsmallind}{\textsc{Block}}
\newcommand{\worldsmallood}{\textsc{Pillars}}
\newcommand{\worldbigind}{\textsc{Tunnels}}
\newcommand{\worldbigood}{\textsc{Chamber}}

\newcommand{\dyn}{\textnormal{Dyn}}

\newcommand{\vu}{{u}}
\newcommand{\vs}{{s}}

\newcommand{\shortsin}[1]{\sin\left(#1\right)}
\newcommand{\shortcos}[1]{\cos\left(#1\right)}
\newcommand{\shorttan}[1]{\tan\left(#1\right)}

\newcommand{\cost}{C(\cdot)}

\newcommand{\methodalilqrastar}{\textsc{AL-iLQR}}

\newcommand{\methodmppi}{\textsc{MPPI}}
\newcommand{\methodflowmppi}{\textsc{FlowMPPI}}

\title{\LARGE \bf
Improving the Resilience of Quadrotors in Underground Environments by Combining Learning-based and Safety Controllers
}

\author{Isaac R. Ward$^{*,1}$, Mark Paral$^{1}$, Kristopher Riordan$^{1}$, and Mykel J. Kochenderfer$^{1}$%
\thanks{$^{*}$Corresponding author: {\tt\small irward@stanford.edu}.}%
\thanks{$^{1}$Stanford Intelligent Systems Laboratory, Department of Aeronautics and Astronautics, Stanford University, Stanford, CA 94305, USA.}%
}

\begin{document}

\maketitle
\thispagestyle{empty}
\pagestyle{empty}

\begin{abstract}


Autonomously controlling quadrotors in large-scale subterranean environments is applicable to many areas such as environmental surveying, mining operations, and search and rescue. Learning-based controllers represent an appealing approach to autonomy, but are known to not generalize well to `out-of-distribution' environments not encountered during training. In this work, we train a normalizing flow-based prior over the environment, which provides a measure of how far out-of-distribution the quadrotor is at any given time. We use this measure as a runtime monitor, allowing us to switch between a learning-based controller and a safe controller when we are sufficiently out-of-distribution. Our methods are benchmarked on a point-to-point navigation task in a simulated 3D cave environment based on real-world point cloud data from the DARPA Subterranean Challenge Final Event Dataset. Our experimental results show that our combined controller simultaneously possesses the liveness of the learning-based controller (completing the task quickly) and the safety of the safety controller (avoiding collision). 



\end{abstract}

\section{Introduction}




Reliable methods for autonomously navigating large-scale, subterranean environments with quadrotors are becoming increasingly relevant, with applications in search and rescue \cite{petrlik2022uavs, bogue2019disaster}, mining operations \cite{papachristos2019autonomous, sikakwe2023mineral}, terrestrial cave exploration \cite{peltzer2022fig,petravcek2021large}, and space exploration \cite{titus2021roadmap,morrell2024robotic,wynne2022fundamental}. 

Learning-based controllers represent an attractive approach to autonomy; they can be learned from data, exhibit extreme maneuverability, and account for non-linear dynamics \cite{kaufmann2023champion}. However, these methods are known to generalize poorly to scenarios not encountered during training \cite{ghosh2021generalization}. Note that we refer to the environment that an algorithm is tuned or trained on as in-distribution (InD); otherwise, it is out-of-distribution (OOD) \cite{guerin2023out}. 


Another branch of potential solutions to autonomy are well-established control theoretic approaches, which can guarantee safety \cite{dawson2020provably}. Such optimal control algorithms allow designers to mathematically optimize desirable performance metrics in the presence of constraints (e.g. collision avoidance and dynamic feasibility). 

There is an inherent tradeoff between safety (something bad will \textit{not} happen, e.g. avoiding colliding into a wall) and liveness (something good \textit{must} happen, e.g. reaching the goal state quickly) \cite{kindler1994safety}. In this work, we balance both considerations by using learning-based controllers when InD, and traditional safety controllers when OOD, relying on a runtime monitor to classify InD from OOD.





{The contributions of this work are threefold}:

\begin{enumerate}

    \item We train a normalizing flow-based optimal control policy (\methodflowmppi{} \cite{power2022variational}) within a Bayesian model-based reinforcement learning paradigm \cite{okada2020variational}. To our knowledge, this is the largest 3D environment on which \methodflowmppi{} has been trained on to date (3D extents of $41\times62\times11$ meters, and an internal volume of $11492$ cubic metres).
    
    \item We design a safety controller that solves for dynamically feasible, obstacle-avoiding trajectories \cite{malyuta2022convex,jacksonilqr}. 

    \item We switch between the two methods at test time using an OOD runtime monitor that considers the environment, current state, and goal state. We experimentally demonstrate that our combined controller simultaneously possesses the liveness of the learning-based controller (completing the task quickly) and the safety of the safety controller (avoiding collisions). 
    
\end{enumerate}

\begin{figure*}[!htb]
    \centering

    
    
    \begin{subfigure}[b]{\textwidth}
        \centering
        \includegraphics[width=0.7\textwidth]{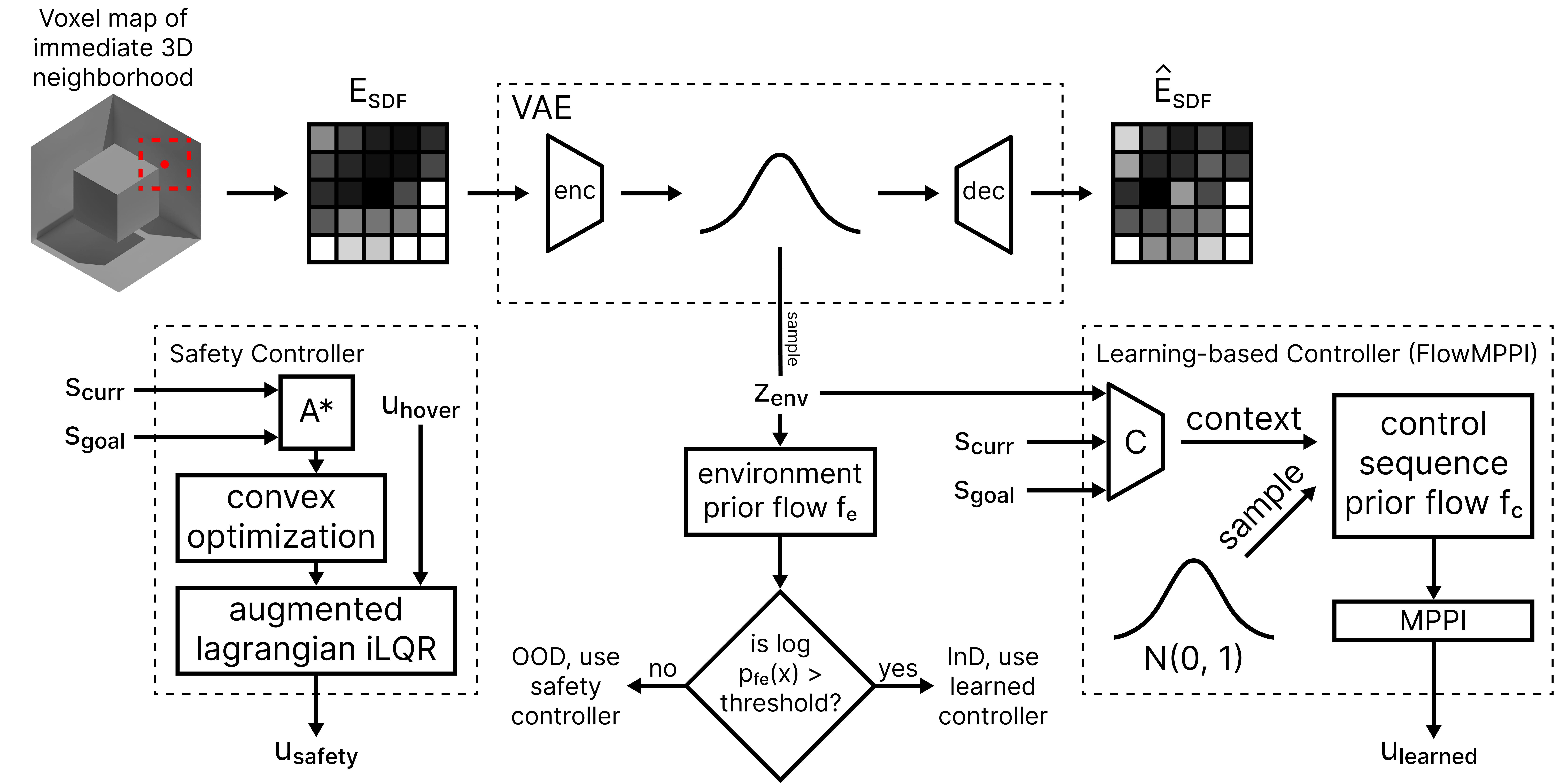}
        \label{fig:overview}
    \end{subfigure}

    \caption{A block diagram explanation of the method introduced in this work. A variational autoencoder is trained to encode signed distance fields of the environment in an online fashion. A prior over the environment encodings is trained in-tandem and probabilistically estimates if any new environmental encoding is from the prior distribution (i.e. `is the current environment similar to what was encountered during training?'). This probability determines which controller is more appropriate to use. The learning-based controller reaches the goal quickly, but is sensitive to out-of-distribution inputs. The safety controller reaches the goal slowly, but is robust to out-of-distribution inputs. By switching between the two, our combined controller achieves superior liveness (reaching the goal quickly) and safety (reaching the goal without collision) properties than either of the constituent methods do on their own.}
    
    \label{fig:two_part_figure}
\end{figure*}

\section{Related Literature}




The learning-based controller in this work is based on \methodflowmppi{} \cite{power2022variational}. This method uses a sampling-based Model Predictive Control (MPC) \cite{morari1999model} framework called Model Predictive Path Integral Control (MPPI) \cite{williams2017model}. MPPI operates by sampling optimal control sequences from a Gaussian prior, simulating them with respect to some known dynamics, assigning a reward to each outcome, and computing an optimal control sequence based on a reward-weighted average of the samples.

The key innovation behind \methodflowmppi{} is the use of a conditional normalizing flow \cite{rezende2015variational} in place of the Gaussian prior when representing the optimal control distribution. Normalizing flows can represent arbitrarily complex distributions, yet can be learned from data. The normalizing flow in \methodflowmppi{} is trained in a Bayesian model-based reinforcement learning paradigm \cite{okada2020variational}, and represents the optimal control distribution. Numerous optimal control samples are drawn from the normalizing flow and a loss is computed based on 1) how goal-directed each sample is, and 2) the breadth of the samples (i.e. exploitation balanced with exploration).

The safety controller used in this work is an Augmented-Lagrangian Iterative Linear Quadratic Regulator \cite{jacksonilqr} (AL-iLQR) tracking the trajectory output from a sequential convex programming (SCP) algorithm. Prior works have demonstrated the value of iLQR-based feedback controllers in minimizing tracking errors \cite{alothman2016quadrotor, sumathy2021adaptive}, and Malyuta et al. \cite{malyuta2022convex} presents a comprehensive guide to generating cost-minimizing, multi-objective, constrained, dynamically feasible, collision avoiding trajectories.

This work also takes advantage of the rich field of OOD detection methods \cite{guerin2023out,yang2024generalized}. Many methods for OOD detection exist, including statistical methods \cite{agia2024unpacking}, comparing latent space representations \cite{ramakrishna2022efficient}, and estimating the likelihood of encodings \cite{gudovskiy2022cflow}, the latter of which is the focus of this work.

We use these methods to create a combined controller than draws on learning-based methods where appropriate, and falls back to conservative SCP methods when needed. Prior works have investigated the construction of meta-controllers that combine 1) non-pedigreed nominal controllers, and 2) recovery controllers, and demonstrated that although safe, recovery controllers may be disruptive to operational objectives \cite{lazarus2020rtsa}.

\begin{figure*}[htbp]
    \centering
    \centering
    \begin{subfigure}[b]{0.46\textwidth}
        \centering
        \includegraphics[width=\textwidth,height=3.2cm,keepaspectratio]{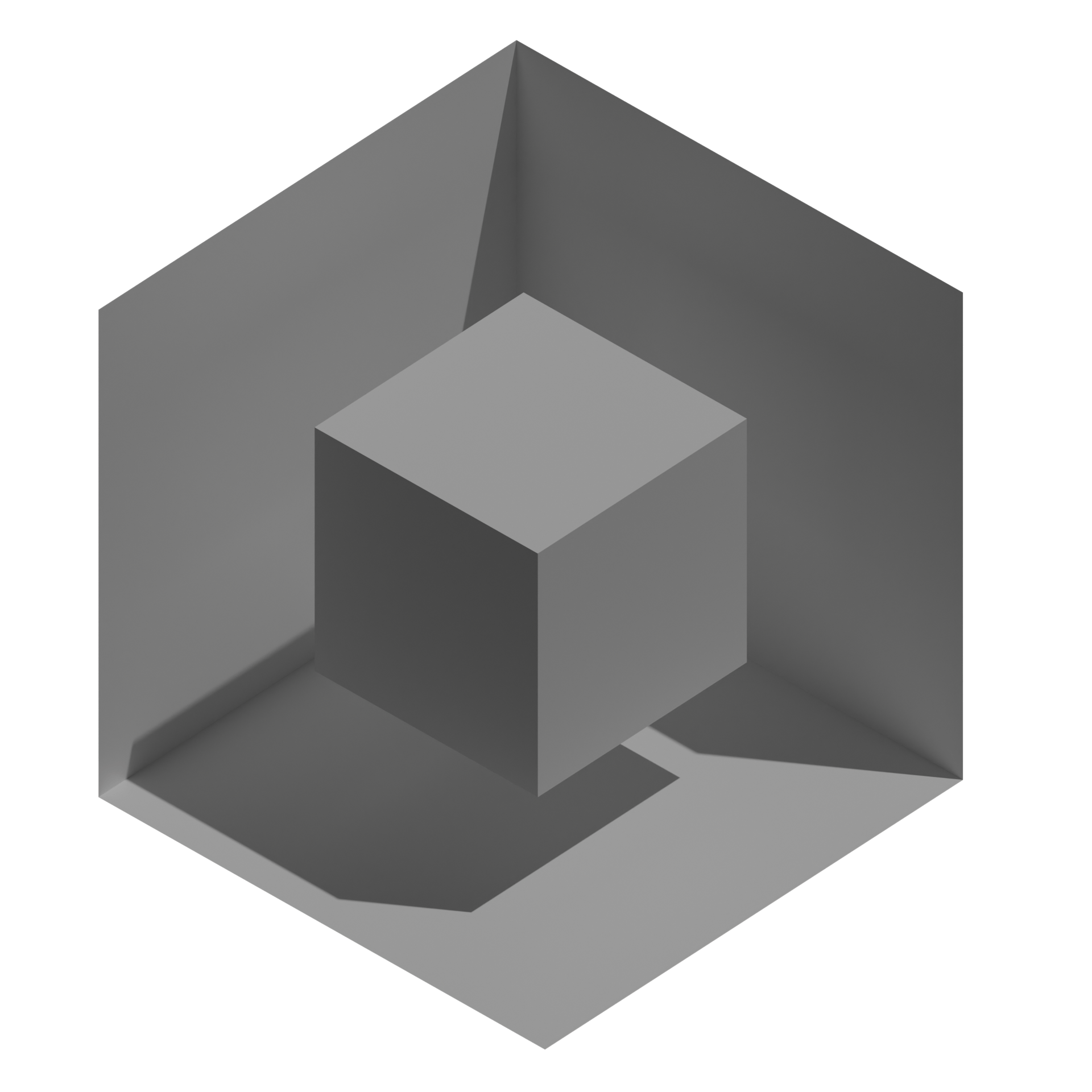}
        \caption{The \worldsmallind{} environment features a block obstacle with $14$ meter side lengths suspended in the center of an enclosed $30\times30\times30$ cubic meter space.}
        \label{fig:block}
    \end{subfigure}
    \hspace{1cm}
    \begin{subfigure}[b]{0.46\textwidth}
        \centering
        \includegraphics[width=\textwidth,height=3.2cm,keepaspectratio]{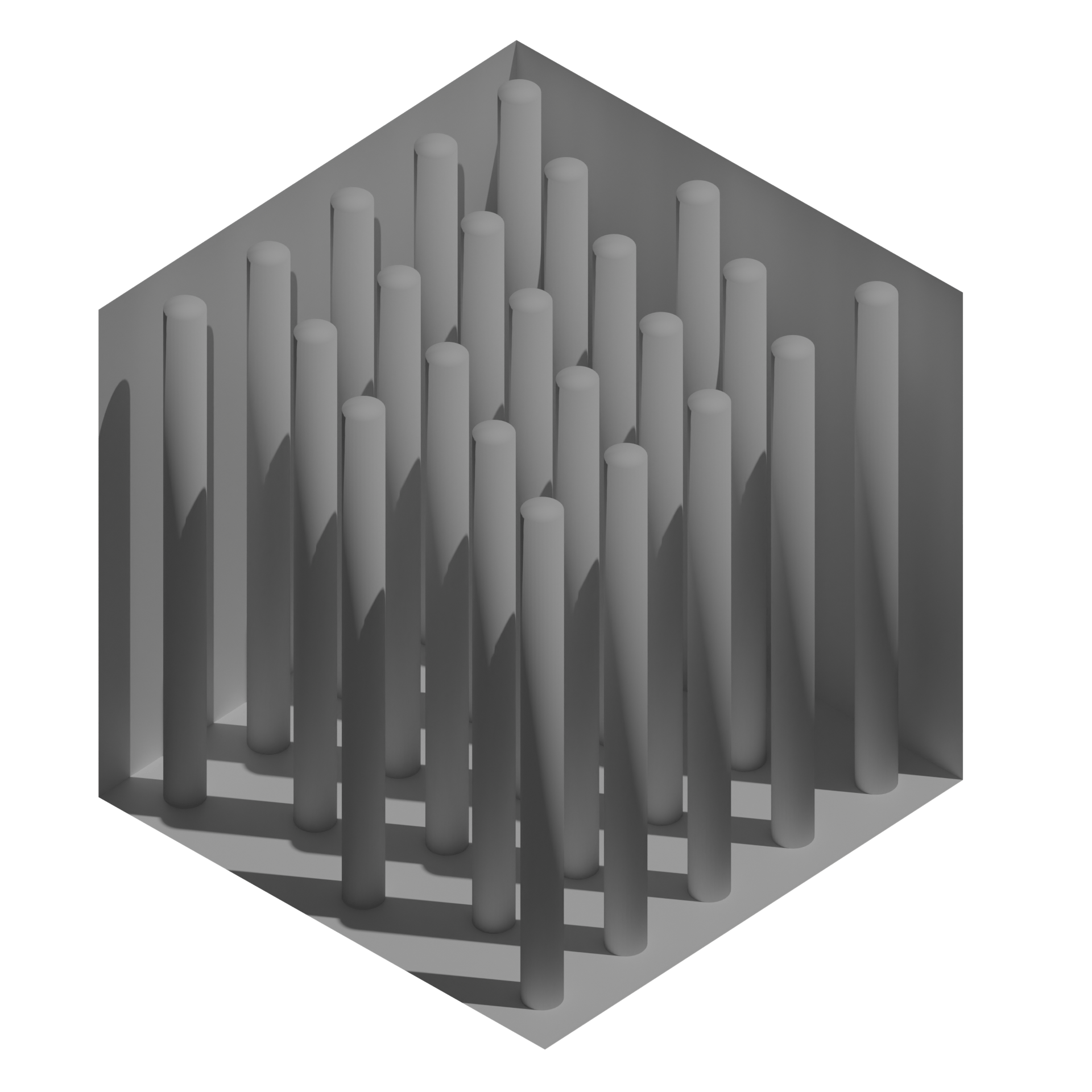}
        \caption{The \worldsmallood{} environment features $23$ cylindrical pillar obstacles (each $2$ meters in diameter) in an enclosed $30\times30\times30$ cubic meter space.}
        \label{fig:pillars}
    \end{subfigure}
    
    \vspace{0.5cm}
    
    \centering
    \begin{subfigure}[b]{0.46\textwidth}
    \centering
    \includegraphics[width=\textwidth,height=3.2cm,keepaspectratio]{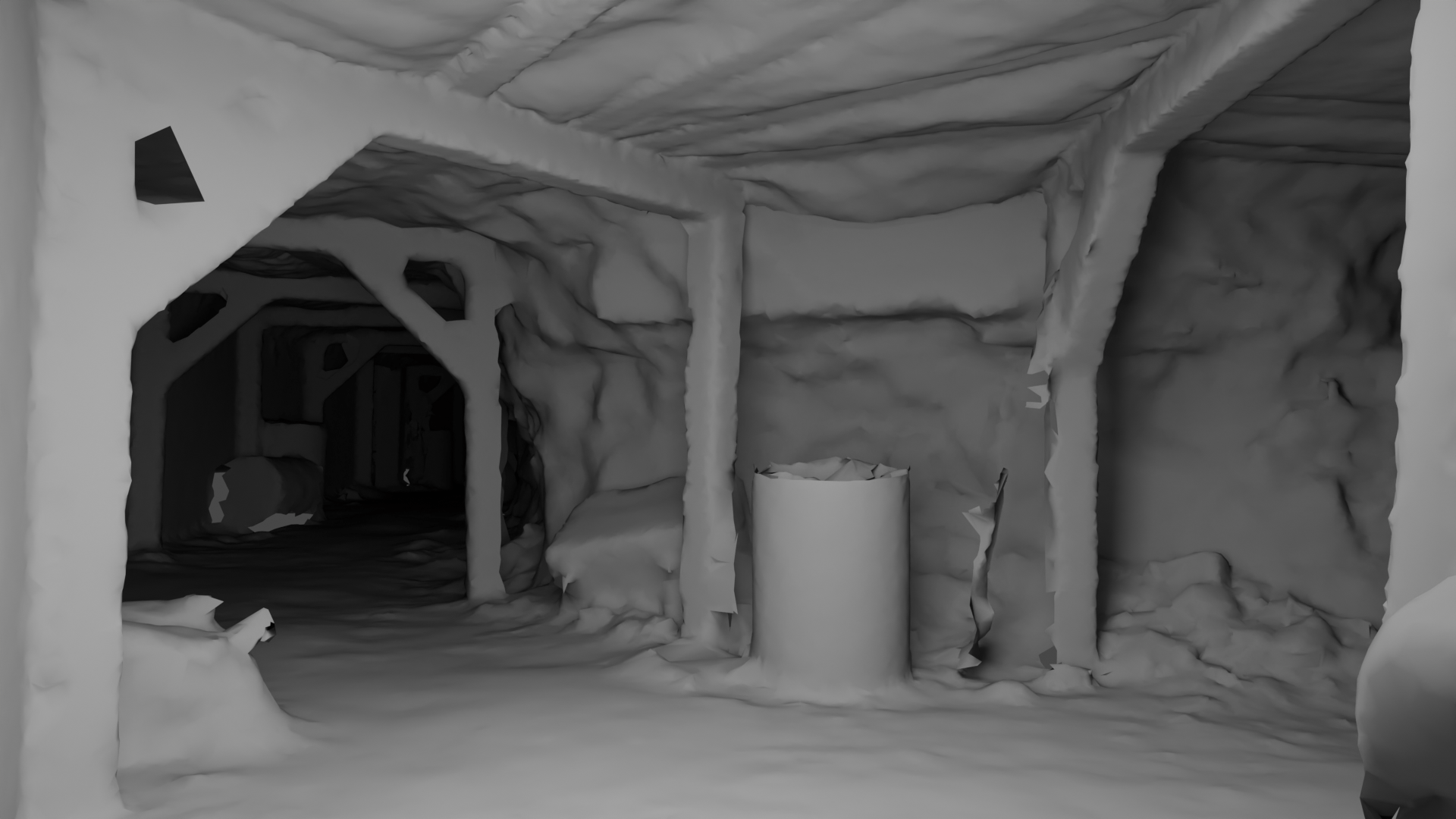}
        \caption{The \worldbigind{} environment features interconnecting mineshafts with cross-sectional areas of approximately $3\times3$ meters and miscellaneous debris (e.g. rubble piles, barrels). The wall surfaces vary between naturally jagged and smooth concrete. This environment is Section~6 of DARPA's Subterranean Challenge Final Event dataset, has extents of $31\times59\times4$ meters, and an internal volume of $536$ cubic meters.}
        \label{fig:tunnels}
    \end{subfigure}
    \hspace{1cm}
    \begin{subfigure}[b]{0.46\textwidth}
        \centering
        \includegraphics[width=\textwidth,height=3.2cm,keepaspectratio]{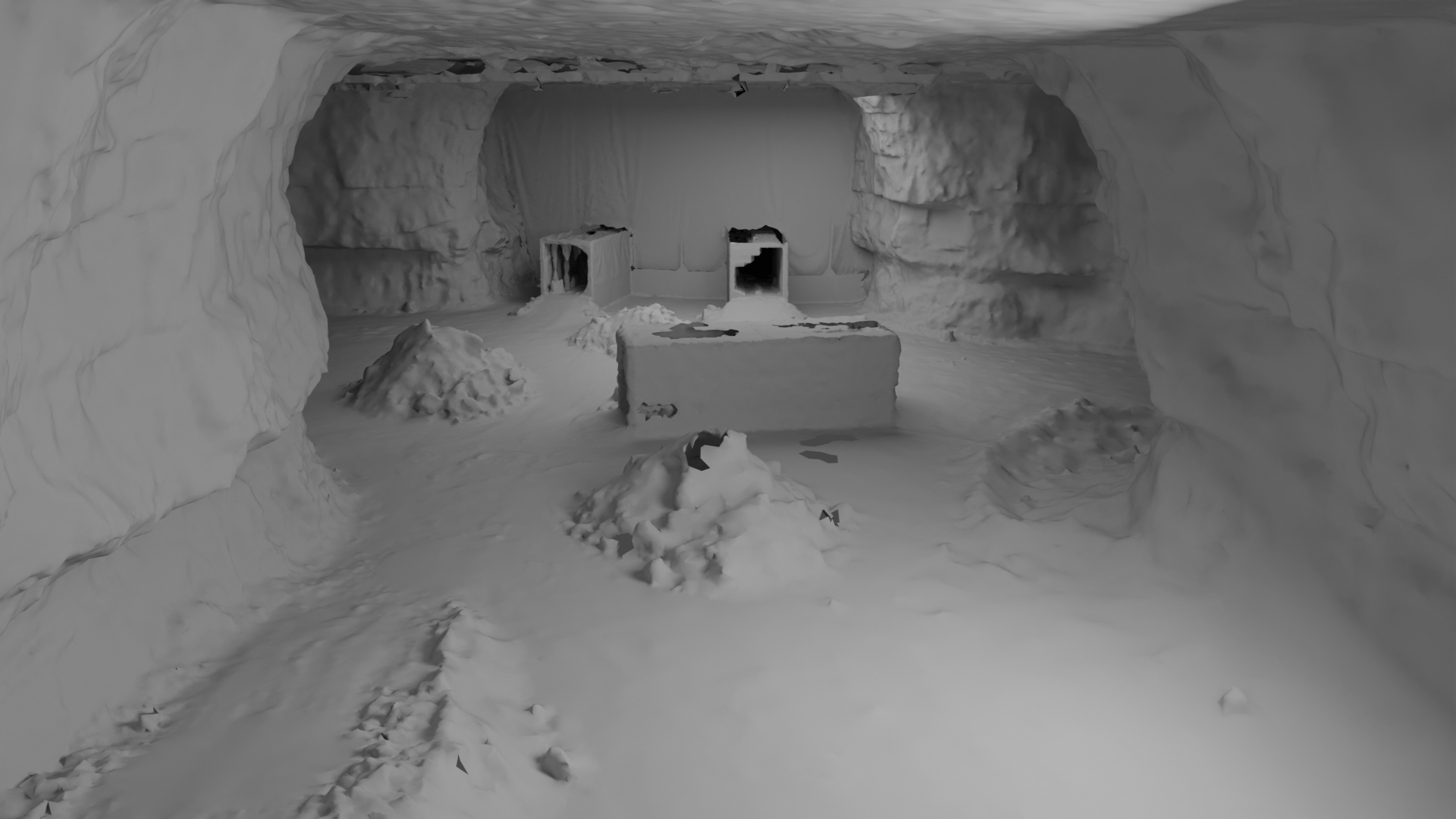}
        \caption{The \worldbigood{} environment is a large cavernous space with piles of rubble, two mineshafts (geometrically similar to the \worldbigind{} environment), and a ramp leading to an elevated area. This environment is Section~4 and Section~5 of DARPA's Subterranean Challenge Final Event dataset, has extents of $41\times62\times11$ meters, and an internal volume of $11492$ cubic metres.}
        \label{fig:chamber}
    \end{subfigure}


    \caption{The four environments used in this work. Each represents a simulated (Figure~\ref{fig:block} and \ref{fig:pillars}) or real-world (Figure~\ref{fig:tunnels} and \ref{fig:chamber}) cave environment. Learning-based methods trained in one environment may not necessarily generalize to different environments during testing, allowing us to test out-of-distribution performance.}
    \label{fig:worlds}
\end{figure*}



\section{Methods}


Our task is simulated point-to-point navigation within enclosed 3D cave environments. We consider two simple handcrafted cave environments and two complex cave environments from DARPA's Subterranean Challenge Final Event dataset, as illustrated in Figure~\ref{fig:worlds}. Start and goal states are selected to be $>25$ meters away from each other, and the quadrotor is initialized at the start position with zero velocity, rotation, and angular velocity. The task is considered complete if the quadrotor stays within one half of its radius to the goal position for at least $3$ seconds. The dynamics that govern the system are defined in Equation~\ref{eq:dynamics}.


\subsection{Model Predictive Path Integral Control (MPPI)}

We implement MPPI (Algorithm~\ref{alg:mppi}) as a baseline, using a multivariate normal distribution as the optimal control distribution as suggested in Williams et al. \cite{williams2017model}. This distribution is updated at each timestep to be centered about the previous optimal control $u^*$, with a covariance matrix that has zero off-diagonal elements. Diagonal elements represent variances, and are computed to be one-quarter of the range of the allowed control inputs. Increasing or decreasing these variances has the effect of widening or tightening the control distribution. 

\subsection{Normalizing Flows with Model Predictive Path Integral Control (\methodflowmppi{})}

We implement a variation of \methodflowmppi{} \cite{power2022variational} and train it within a Bayesian model-based reinforcement learning paradigm \cite{okada2020variational}. The implementation follows Algorithm~\ref{alg:mppi} with a conditional normalizing flow representing the optimal control distribution. 

This flow is conditioned on the task variables (the start state and goal state), and the encoding of the quadrator's immediate environment (using a variational autoencoder \cite{kingma2013auto}). We refer to the concatenated task variables and environment encoding as the context vector $C$. In practice, this ensures that the samples drawn from the optimal control distribution are contextually informed; they are goal directed (move from the start state to the goal state) and avoid collisions (by accounting for the immediate environment).



\begin{algorithm}[h]
\caption{Model Predictive Path Integral Control}
\label{alg:mppi}
\begin{algorithmic}[1]
\algsetup{linenosize=\scriptsize} 
\algsetup{linenodelimiter=} 

\REQUIRE initial state $\vs_1$, samples $K$, horizon $H$, \\forward dynamics function $\dyn$,\\ optimal control distribution $\mathcal{P}^*$,\\ cost function $\cost$, weighting sharpness parameter $\lambda$.

\STATE // Store state and control trajectories and their costs
\STATE $\tau_{\vec{s}} \leftarrow [ \; ]$; $\tau_{\vec{u}} \leftarrow [ \; ]$; $C \leftarrow [ \; ]$
\FOR{$k = 1$ to $K$}

    \STATE $\tau_{\vec{s}}^k \leftarrow [ \vs_{\textnormal{1}} ]$; $\tau_{\vec{u}}^k \leftarrow [ \; ]$

    \FOR{$h = 1$ to $H$}
        
        \STATE // Execute the sampled control and get the next state
        \STATE $\vu_h^k \leftarrow \textnormal{sample}(\mathcal{P}^*)$
        \STATE $\vs_{h}^{\:k} \leftarrow \dyn(\tau_{{s}}^k[-1], \vu_h^{\:k})$

        \STATE // Append the state and control to the trajectories
        \STATE $\textnormal{append}(\tau_{{s}}^k, \vs_{h}^{\:k})$; $\textnormal{append}(\tau_{{u}}^k, \vu_h^{\:k})$
    
    \ENDFOR 
    \STATE // Store state and control trajectory k
    \STATE $\textnormal{append}(\tau_{{s}}, \tau_{{s}}^k)$; $\textnormal{append}(\tau_{{u}}, \tau_{\vec{u}}^k)$
    \STATE // Relate the trajectory to a cost and store
    \STATE $\textnormal{append}(C, \sum_{h=1}^H C(\tau_{{s}}^k[h], \tau_{{u}}^k[h]) )$
\ENDFOR
\STATE // Compute weights for each trajectory based on costs
\STATE $w \leftarrow \text{softmax}(-\lambda \cdot C)$
\STATE // Update the optimal control distribution (as needed)
\STATE $\mathcal{P}^* \leftarrow \textnormal{update}(\cdot)$
\STATE // Combine optimal cost-weighted control sequence,
\STATE // and return the immediate (first) optimal control 
\STATE $u^* \leftarrow (\sum_{k=1}^{K} w_k \tau_{{u}}^k)[1]$
\STATE return $u^*$

\end{algorithmic}
\end{algorithm}

\subsection{Sequential Convex Programming for Trajectory Optimization}


We determine non-colliding, dynamically feasible trajectories using sequential convex programming (SCP). The method iteratively refines an initial trajectory by solving a sequence of convex optimization problems \cite{malyuta2022convex}.

We begin by computing a volume that 1) contains the start state $s_{\text{init}}$ and goal state $s_{\text{goal}}$, and 2) contains no environmental obstacles. We compute this volume as a set of $V$ spheres by first solving for the shortest path between $s_{\text{init}}$ and $s_{\text{goal}}$ using the A$^*$ search algorithm \cite{hart1968formal}. We dilate the voxel map representation of the environment by the quadrotor's radius to ensure that there would be no collision even if the quadrotor were centered at any point along the resulting path. 

The first sphere is set to be centered at $s_{\text{init}}$ and is iteratively grown from a zero radius until it intersects with the environment. The next sphere is centered at the point that is both on the A$^*$ path and on the previous sphere's surface. Again we grow the sphere until collision, and this process repeats until a sphere contains $s_{\text{goal}}$. The union of these $V$ spheres is a contiguous volume that contains $s_{\text{init}}$ and $s_{\text{goal}}$. 

We initialize the solver with a state trajectory of shape $(T,12)$, where the positional elements (first three elements of the 12-dimensional state vector) are the A$^*$ path solutions, and all other values (velocity, rotation, and angular velocity) are zero. This initial state trajectory is dynamically infeasible but will be iteratively improved upon. We initialize the action trajectory with a set of controls that causes the quadcopter to hover in place. We find that this trajectory initialization warm-starts the solver and leads to faster solve times \cite{malyuta2022convex}.


The optimization is a multiobjective cost function that simultaneously minimizes deviation from the goal, control effort, navigation slack (staying within the non-collision volume $V$), and dynamics slack (discussed further below):
\begin{align}
\label{eq:obj}
J = &\ w_{\text{dist}} \sum_{t=1}^{T} \lvert s_t - s_{\text{goal}} \rvert^2 \;+\; w_{\text{ctrl}} \sum_{t=1}^{T-1} \lvert u_t \rvert^2 \;+ \nonumber\\ 
&\ w_{\text{nav}} \sum_{t=1}^{T} \sum_{v=1}^{V} \xi_{t,v}^{\text{nav}} \;+\; w_{\text{dyn}} \max \lvert \xi^{\text{dyn}} \rvert
\end{align}
%
The trajectory is constrained to start and end at fixed states
\begin{equation}
    s_1 = s_{\text{init}} \quad \text{and} \quad s_T = s_{\text{goal}}
\end{equation}
Control inputs are constrained to dynamically feasible limits
\begin{equation}
u_t \in [u_{\min}, u_{\max}], \; \forall t \in [1, T-1]
\end{equation}
Successive states are constrained to satisfy the affinized dynamics with some allowed slack at each timestep $\xi^{\text{dyn}}$
\begin{equation}
s_{t+1} = A_t s_t + B_t u_t + C_t + \xi_t^{\text{dyn}}, \; \forall t \in [1, T-1]
\end{equation}
$A_t$, $B_t$, and $C_t$ represent the dynamics affinized about the previous iteration's states $s_t$ and controls $u_t$. To maintain the accuracy of this affine approximation we enforce a trust region
\begin{equation}
\lvert s_t - s_t^{\text{prev}} \rvert \leq r_s \; \forall t \in [1, T]
\end{equation}
\begin{equation}
\lvert u_t - u_t^{\text{prev}} \rvert \leq r_u, \; \forall t \in [1, T-1]
\end{equation}
and the maximum dynamics slack must stay below some defined threshold $\xi^{\text{dyn}}_{\max}$
\begin{equation}
\max \lvert \xi_t^{\text{dyn}} \rvert \leq \xi^{\text{dyn}}_{\max}
\end{equation}
%
%
Note that the trust region and maximum allowed dynamics slack are geometrically decayed at each iteration according to a decay schedule (Table~\ref{tab:hyperparams}).
%
We define a function $D(\cdot)$ which takes a state trajectory and returns a $(T,V)$-shaped matrix where the element $t,v$ represents the signed distance field value of the $t^{th}$ state's position with respect to the $v^{th}$ collision-free sphere \cite{oleynikova2016signed}. We also introduce a navigation slack variable $\xi^{\text{nav}}$ which is solved for at each iteration, and represents how much we deviated from the non-collision volume in that iteration. We introduce the first navigation constraint over every element of the navigation slack:
\begin{equation}
\xi^{\text{nav}}_t \leq D(s_t), \; \forall t \in [1, T]
\end{equation}
which states that the navigation slack $\xi^{\text{nav}}$ in this iteration must be less than the true signed distance field values from the current trajectory solution. We then introduce the second navigation constraint:
\begin{equation}
\text{diag} \left( \frac{e^{\sigma D(s^{\text{prev}})}}{\sum e^{\sigma D(s^{\text{prev}})}} (\xi^{\text{nav}} - D(s^{\text{prev}}))^T \right) + \frac{\log \sum e^{\sigma D(s^{\text{prev}})}}{\sigma}  \geq 0
\end{equation}
which uses a softmax with temperature $\sigma$ to ensure that each point in the trajectory is moving more towards the center of a non-colliding sphere in successive iterations \cite{malyuta2022convex}.

This problem is solved iteratively until the solution does not meaningfully improve with respect the objective $J$ ({Equation}~\ref{eq:obj}). 



\subsection{Augmented Lagrangian Iterative Linear Quadratic Regulator (AL-iLQR)}



The Iterative Linear Quadratic Regulator (iLQR) is an optimization framework which leverages the Linear Quadratic Regulator (LQR) control policy and applies it to systems with nonlinear dynamics, using iterative dynamics linearization about the nominal trajectory $s^{\textnormal{nom}}$, $u^{\textnormal{nom}}$. The optimal controller follows the control law:
\begin{equation}
    u_t=u_t^{\textnormal{nom}}+K_t (s_t-s_t^{\textnormal{nom}})+d_t
    \label{eq:iLQRcontrol}
\end{equation}
where $K_t$ is the feedback gain, $d_t$ is the feedforward gain, $s_t$ is the state for timestep $t\in [1,T]$, and $u_t$ is the control input for timestep $t\in [1,T-1]$.

There are many practical constraints for a quadrotor controller that must be enforced, such as control bounds on rotor inputs. Traditional iLQR does not allow for this, so we use Augmented Lagrangrians (AL), which reposes a standard optimization problem with explicit constraints into a problem with soft constraints enforced in the cost function \cite{jacksonilqr}. 
The new optimization cost after applying the AL technique is referred to as the Augmented Lagrangian $\mathcal{L}_A$, and includes a Lagrange multiplier $\lambda$ and penalty matrix $I_{\mu}$. 
Using AL within the iLQR tracking framework results in the AL-iLQR tracking problem:
\begin{mini!}|s|[2]<b>{{u}_{1:T-1}}{\mathcal{L}_A(s_{1:T},u_{1:T-1})}{\label{eq:aliLQRoptpost}}{}
\addConstraint{s_{t+1}}{= A_t s_t + B_t u_t, \quad \forall t = [1, T-1]}
\end{mini!}
which states that a control sequence that minimizes the AL: 
\begin{equation}
\resizebox{0.48\textwidth}{!}{$
\begin{aligned}
    &\mathcal{L}_A(s_{1:T},u_{1:T-1}) 
    = \ell_{T} + \left( \lambda_{T}^\top + \frac{1}{2} c_{T}^\top I_{\mu, {T}} \right) c_T
    &+ \sum_{t=1}^{T-1} \left(\ell_t + \left( \lambda_t^\top + \frac{1}{2} c_t^\top I_{\mu, t} \right) c_t\right)
\end{aligned}
$}
\label{eq:augmentedlagrangian}
\end{equation}
whilst maintaining dynamic feasibility, must be found. We further define bounds on the control inputs
\begin{equation}
\begin{aligned}
    \left[c_t(s_t,u_t),c_{T}(s_{T})\right] &= 
        \left[
        \begin{bmatrix}
        u_t-u_{\textnormal{max}} \\
        u_{\textnormal{min}}-u_t
        \end{bmatrix}, \; 0
    \right]
\end{aligned}
\label{eq:constraints}
\end{equation}
and define the standard iLQR tracking costs, which penalize divergence from the desired state trajectory $s^{\textnormal{track}}$ using state cost, control cost, and terminal state cost matrices $Q$, $R$, and $Q_T$: 
\begin{equation}
\begin{aligned}
    \ell_t(s_t, u_t) &= (s_{t} - s_t^{\textnormal{track}})^\top Q (s_{t} - s_t^{\textnormal{track}}) + u_{t}^\top R u_{t}\\
    \ell_{T}(s_{T}) &= \frac{1}{2} (s_{T} - s_{T}^{\textnormal{track}})^\top Q_{T} (s_{T} - s_{T}^{\textnormal{track}})
\end{aligned}
\label{eq:ilqr_cost}
\end{equation}
Which are minimized.
\section{Results}


\begin{table*}[!htbp]
    \centering
    \renewcommand{\arraystretch}{1.2}
    \setlength{\tabcolsep}{6pt}  
    \caption{Test results across all environments. Methods are trained and tuned on the in-distribution environment before evaluation on the corresponding out-of-distribution environment environment (meaning that we train on \worldsmallind{} before evaluating on \worldsmallind{} and \worldsmallood{}, and we train on \worldbigind{} before evaluating on \worldbigind{} and \worldbigood{}). We consider cases where the training environment matches the test environment to be in-distribution, and cases where the training environment does not match the test environment to be out-of-distribution. $N=100$ randomized start and end goals are pre-generated for each environment so that each method can be tested consistently. \textbf{Abbreviations}: success rate SR, total time to complete task $\bar{T}_{\text{done}}$, average quadrotor velocity $\bar{v}$, average flight trajectory length $\bar{d}$, and average total control effort throughout trajectory $\bar{u}$. 
    }

    
    \begin{tabular}{l ccccc ccccc}
        \toprule
         & \multicolumn{5}{c}{\worldsmallind{} (small, in-distribution) } & \multicolumn{5}{c}{\worldsmallood{} (small, out-of-distribution) } \\
         
        \cmidrule(lr){2-6} \cmidrule(lr){7-11}  
        
        & SR & $\bar{T}_{\text{done}}$ & $\bar{v}$ & $\bar{d}$ & $\bar{u}$ 
        & SR & $\bar{T}_{\text{done}}$ & $\bar{v}$ & $\bar{d}$ & $\bar{u}$ \\ 
        Methods & (\%) & (s) & (ms$^{-1}$) & (m) & (N)
        & (\%) & (s) & (ms$^{-1}$) & (m) & (N) \\ 
        \midrule
        \methodalilqrastar{}     & 100 & 40.32 & 2.24 & 43.22 & 2.62 & 94 & 60.29 & 2.85 & 54.39 & 3.27 \\
        \methodmppi{}            & 98 & 34.58 & 5.03 & 58.50 & 4.19 & 93 & 37.10 & 5.47 & 42.99 & 4.02 \\
        \methodflowmppi{}       & 100 & 36.15 & 4.98 & 54.57 & 4.76 & 71 & 28.16 & 4.70 & 35.32 & 4.96 \\
        Combined (ours)            & 99 & 39.33 & 4.74 & 51.22 & 4.83 & 92 & 51.18 & 3.01 & 49.68 & 3.99 \\
        
        \midrule
        
         & \multicolumn{5}{c}{\worldbigind{} (large, in-distribution) } & \multicolumn{5}{c}{\worldbigood{} (large, out-of-distribution)  } \\
         
        \cmidrule(lr){2-6} \cmidrule(lr){7-11}  
        
        \methodalilqrastar{}     & 93 & 144.81 & 2.18 & 64.12 & 2.14 & 86 & 133.47 & 2.06 & 39.97 & 2.63 \\ 
        \methodmppi{}            & 90 & 45.60 & 3.54 & 91.97 & 5.62  & 78 & 40.54 & 6.59 & 53.96 & 4.14 \\ 
        \methodflowmppi{}        &  88 & 43.11 & 3.89 & 78.38 & 5.11 & 76 & 38.43 & 6.04 & 48.13 & 4.91 \\ 
        Combined (ours)             & 92 & 46.44 & 3.75 & 80.20 & 5.02 & 84 & 50.52 & 5.11 & 45.76 & 4.22 \\ 
        \bottomrule
    \end{tabular}
    \label{tab:oodenvsresults}
\end{table*}



\subsection{The Learning-based Controller Completes the Task Faster but is More Sensitive to OOD Scenarios }

In Table~\ref{tab:oodenvsresults} we observe that \methodflowmppi{} completes tasks the fastest out of all our controllers; in all environments apart from \worldsmallind{}, \methodflowmppi{} has the fastest task completion time for successful tasks, demonstrating the value of model-based reinforcement learning for InD inputs.

However, we observe a large drop in success rate when going from InD to OOD when using \methodflowmppi{}. In the small environments, success rate drops from $100\%$ to $71\%$, and in the large environments, success rate drops from $93\%$ to $76\%$, illustrating the method's OOD-sensitivity. We note that the failure of learning-based methods to generalize to new environments is well-observed \cite{ghosh2021generalization}.

\subsection{The Safety Controller Completes the Task Slower but is More Robust to OOD Scenarios}

Compared to the learning-based controller, the safety controller completes tasks much slower; in all four of the tested environments, the safety controller has the slowest completion times by far (see Table~\ref{tab:oodenvsresults}). However, the controller exhibits a less severe drop in success rate compared to \methodflowmppi{} when going from InD to OOD. In the small environments, success rate drops from $100\%$ to $94\%$, and in the large environments, success rate drops from $88\%$ to $86\%$. We also note that overall the safety controller produces smoother, lower control effort, shorter trajectories. We expect that this is due to SCP globally optimizing the entire path, whereas \methodflowmppi{} and \methodmppi{} are myopic due to their local horizons.

These results experimentally show that the safety controller is more resilient to OOD than the learning-based controller, despite being much slower. This illustrates the tradeoff between liveness and safety properties \cite{kindler1994safety}.

\subsection{Combining both Controllers with an OOD Detector Improves Task Completion Speed and OOD Resilience }


Switching between the learning-based and safety controller based on an OOD score results in a controller that exhibits the best properties of its constituents. In the small environments (\worldsmallind{} and \worldsmallood{}) the combined controller achieves comparable success rate (within $1\%$) to \methodalilqrastar{}, yet the average task completion speed is greatly reduced on account of \methodflowmppi{} being used while InD (see Table~\ref{tab:oodenvsresults}). 

We note that the combined method having a lower success rate than the safety controller suggests that either 1) the OOD scenarios may not be perfectly classified, or 2) that the safety controller has a higher failure rate when switched to. Both scenarios are plausible; OOD classification is imperfect and the combined controller often switches to the safety controller when in the presence of close obstacles, or dynamically challenging states (i.e. states not encountered during training).


Table~\ref{tab:oodenvsresults} shows a similar pattern on the large environments (\worldbigind{} and \worldbigood{}). The combined controller achieves comparable success rates to the safety controller (within $2\%$), yet greatly reduced completion times due to the InD performance of the learning-based constituent controller.

\section{Conclusion}

In this work we have used out-of-distribution detection to determine when a learning-based controller and a safety controller should be used in the context of a point-to-point quadrotor navigation task in an underground 3D environment.

We find that learning-based methods (specifically, FlowMPPI) complete tasks quickly when in-distribution, but exhibit lower success rates when out-of-distribution. We find that our safety controller (based on sequential convex programming and AL-iLQR) is more robust to out-of-distribution scenarios, but exhibits overall slower task completion. By combining both, we create a controller that completes task quickly when in-distribution and maintains safety when out-of-distribution.



\section{Acknowledgments}

Toyota Research Institute (TRI) provided funds to assist the
authors with their research, but this article solely reflects the
opinions and conclusions of its authors and not TRI or any
other Toyota entity.

\bibliographystyle{IEEEtran}
\bibliography{IEEEabrv,references}

\section*{Appendix}

\subsection{Dynamics}

\label{s:dynamics}

The system has a $12$-dimensional state-space representing the 3D position, velocity, rotation (Euler angles), and angular velocity, and a $4$-dimensional control-space representing the angular velocities of the rotors. Our discretized dynamics model follows the implementation outlined in \cite{sabatino2015quadrotor}, and assumes negligible motor resistance (i.e. motor power is assumed to be proportional to angular velocity), stationary air, linear fluid friction, negligible rotational drag, and does not account for blade flapping:

%
\begin{equation}
    \tiny
    \label{eq:dynamics}
    \hspace{-2mm}\begin{bmatrix}
        x \\ y \\ z \\ r_z \\ r_y \\ r_x \\ \dot{x} \\ \dot{y} \\ \dot{z} \\ \omega_x \\ \omega_y \\ \omega_z
    \end{bmatrix}_{t+1}\hspace{-2mm}
    =
    \begin{bmatrix}
        x \\ y \\ z \\ r_z \\ r_y \\ r_x \\ \dot{x} \\ \dot{y} \\ \dot{z} \\ \omega_x \\ \omega_y \\ \omega_z
    \end{bmatrix}_{t}\hspace{-1mm}
    +
    \Delta t
    \begin{bmatrix}
        \dot{x} \\
        \dot{y} \\
        \dot{z} \\
        \omega_y \frac{\shortsin{r_x}}{\shortcos{r_y}} + \omega_z \frac{\shortcos{r_x}}{\shortcos{r_y}} \\
        \omega_y \shortcos{r_x} - \omega_z \shortsin{r_x} \\
        \omega_x + \omega_y \shortsin{r_x} \shorttan{r_y} + \omega_z \shortcos{r_x} \shorttan{r_y} \\
        -\left(\shortsin{r_x} \shortsin{r_z} + \shortcos{r_z} \shortcos{r_x} \shortsin{r_y}\right) k\frac{\sum_{i=1}^{4} \omega_i}{m} \\
        -\left(\shortcos{r_x} \shortsin{r_z} \shortsin{r_y} - \shortcos{r_z} \shortsin{r_x}\right) k\frac{\sum_{i=1}^{4} \omega_i}{m} \\
        g - \shortcos{r_x} \shortcos{r_y} k\frac{\sum_{i=1}^{4} \omega_i}{m} \\
        \frac{(I_y - I_z) \omega_y \omega_z + k r (\omega_3^2 - \omega_1^2)}{I_x} \\
        \frac{(I_z - I_x) \omega_x \omega_z + k r (\omega_4^2 - \omega_2^2)}{I_y} \\
        \frac{(I_x - I_y) \omega_x \omega_y + b ((\omega_2^2 + \omega_4^2) - (\omega_1^2 + \omega_3^2))}{I_z} 
    \end{bmatrix}_{t}
\end{equation}
%

\vspace{2mm}
where $(x,y,z)$ is the position, $(\dot{x},\dot{y},\dot{z})$ is the velocity, $(r_z,r_y,r_x)$ is the 3D rotation (Euler angles), $(\omega_x,\omega_y,\omega_z)$ are the body's angular velocities, $(\omega_1,\omega_2,\omega_3,\omega_4)$ are the rotor's angular velocities, $k$ is the thrust coefficient, $m$ is the quadrotor's mass, $b$ is the yaw-drag coefficient, $r$ is the quadrotor's radius, and $(I_x,I_y,I_z)$ are the quadrotor's moments of inertia about the $(\hat{x},\hat{y},\hat{z})$ axes. The gravitational acceleration is $g=9.81$m/s$^2$, and the time step is $\Delta t=0.025$s. Control bounds are $u_{\min} = 0$ and $u_{\max} = 4$.



\subsection{Reproducibility}
\label{s:reprod}


All experiments were executed on a workstation with an AMD Ryzen Threadripper 3970X 32-Core 64-Thread Processor (3.69 GHz) CPU and 128 GB of usable RAM. The training of neural networks was executed on NVIDIA GeForce RTX 3090 GPUs using PyTorch Lightning. The code and data required to reproduce these results is available at \url{https://github.com/sisl/underground}. The hyperparameters used in our algorithms are specified in Table~\ref{tab:hyperparams}.

\newpage
\begin{table}[!ht]
    \centering
    \caption{Descriptions of hyperparameters and their values as used in this work. }
    \begin{tabular}{l c}
        \toprule
        
        \multicolumn{2}{c}{Sequential Convex Programming Hyperparameters} \\
        \midrule
        
        Initial state trust region radius $r_s$ & $1$ \\
        State trust region radius decay $\gamma_{s}$ & $0.95$ \\
        Initial control trust region radius $r_c$ & $1$ \\
        Control trust region radius decay $\gamma_{c}$ & $0.95$ \\
        Maximum dynamics slack $\xi^{\text{dyn}}_{\max}$ & $1$ \\
        Maximum dynamics slack decay $\gamma_{\text{dyn}}$ & $0.5$ \\
        Navigation sigma $\sigma$ & $50$ \\
        Minimum $J$ improvement & $0.01$ \\
        Goal directedness term weight $w_{\text{dist}}$ & $0.1$ \\
        Control effort term weight $w_{\text{ctrl}}$ & $0.1$ \\
        Navigation slack term weight $w_{\text{nav}}$ & $0.01$ \\
        Dynamics slack term weight $w_{\text{dyn}}$ & $100$ \\
        


        \midrule
        \multicolumn{2}{c}{\methodalilqrastar{} hyperparameters} \\
        \midrule
        
        State cost matrix $Q_k$ & $Q_{k,ij} = \scriptstyle
            \begin{cases} 
            5, & i = j \;\wedge\; i \leq 3 \\ 
            1, & i = j \;\wedge\; i > 3 \\ 
            0, & i \neq j 
            \end{cases}$ \\

        Final state cost matrix $Q_K$ & $Q_{K,ij} = \scriptstyle
            \begin{cases} 
            100, & i = j \;\wedge\; i \leq 3 \\ 
            10, & i = j \;\wedge\; i > 3 \\ 
            0, & i \neq j 
            \end{cases}$ \\
        
        Action cost matrix $R_k$ & $I_4$ \\

        Maximum \# of path segments & $3$ \\

        \# points per meter resampled to & $10$ \\

            
        \midrule
        \multicolumn{2}{c}{\methodmppi{} hyperparameters} \\
        \midrule
        
        Samples K & $1024$ \\
        Horizon H & $50$ \\
        Temperature $\lambda$ & $20$ \\

            
        \midrule
        \multicolumn{2}{c}{\methodflowmppi{} hyperparameters} \\
        \midrule

        Samples K & $1024$ \\
        Horizon H & $50$ \\
        Temperature $\lambda$ & $20$ \\
        Neighborhood size & $6\times6\times6$ m$^3$ \\
        VAE input size & $864$ \\
        VAE latent size & $128$ \\
        CNF \# flow layers & $8$ \\
        CNF hidden layers' size & $256$ \\
        CNF hidden layer type & AffineCouplingTransform \\
        Context vector input size & $152$ \\
        Context vector embedding size & $64$ \\
        \# training episodes & $256$ \\

        
        \midrule
        \multicolumn{2}{c}{Combined method's hyperparameters} \\
        \midrule
        
        Normalized OOD threshold (log space) & 3.12 \\
        
        \bottomrule
    \end{tabular}
    \label{tab:hyperparams}
\end{table}

\end{document}